\documentclass{article}

\usepackage{cite}      % for numeric citations
\usepackage{PRIMEarxiv}
\usepackage[table]{xcolor}% http://ctan.org/pkg/xcolor
\usepackage[utf8]{inputenc} % allow utf-8 input
\usepackage[T1]{fontenc}    % use 8-bit T1 fonts
\usepackage{hyperref}       % hyperlinks
\usepackage{url}            % simple URL typesetting
\usepackage{booktabs}       % professional-quality tables
\usepackage{amsfonts}       % blackboard math symbols
\usepackage{nicefrac}       % compact symbols for 1/2, etc.
\usepackage{microtype}      % microtypography
\usepackage{lipsum}
\usepackage{fancyhdr}       % header
\usepackage{graphicx}       % graphics
\graphicspath{{media/}}     % organize your images and other figures under media/ folder
\title{Exploring the Effectiveness of GPT Models in Test-Taking: A Case Study of the Driver's License Knowledge Test
}
\author{
  Saba Rahimi, Tucker Balch, Manuela Veloso \\
  J. P. Morgan AI Research, New York, NY, USA \\
  \texttt\ saba.rahimi@jpmorgan.com \\
}
\begin{document}
\maketitle
\begin{abstract}
Large language models such as Open AI’s Generative Pre-trained Transformer (GPT) models are proficient at answering questions, but their knowledge is confined to the information present in their training data. This limitation renders them ineffective when confronted with questions about recent developments or non-public documents. Our research proposes a method that enables GPT models to answer questions by employing context from an information source not previously included in their training data. The methodology includes preprocessing of contextual information, the embedding of contexts and queries, constructing prompt through the integration of context embeddings, and generating answers using GPT models. We applied this method in a controlled test scenario using the California Driver's Handbook as the information source. The GPT-3 model achieved a 96\% passing score on a set of 50 sample driving knowledge test questions. In contrast, without context, the model's passing score fell to 82\%. However, the model still fails to answer some questions correctly even with providing library of context, highlighting room for improvement. The research also examined the impact of prompt length and context format, on the model’s performance. Overall, the study provides insights into the limitations and potential improvements for GPT models in question-answering tasks.
\end{abstract}

% keywords can be removed

\section{Introduction}

In recent years, the advent of large-scale language models, such as GPT models \cite{brown2020language}, \cite{chen2021evaluating}, has revolutionized natural language understanding and generation \cite{rae2021scaling}, \cite{chowdhery2022palm}, \cite{thoppilan2022lamda}, \cite{zhang2022opt}, \cite{smith2022using}, \cite{fedus2022switch}. These models have demonstrated exceptional performance in a wide range of tasks, including machine translation, summarizing, and question-answering. Their prowess stems from their pre-training on vast amounts of text data, enabling them to learn complex language patterns and structures. As a result, these models possess a remarkable ability to generate human-like responses to a variety of prompts. Prior research has explored the capabilities of GPT models in different question-answering domains. For instance, studies have assessed the performance of these models in answering questions from the bar exam \cite{bommarito2022gpt}, \cite{katz2023gpt}, medical tests \cite{lievin2022can}, \cite{kung2023performance}, \cite{nori2023capabilities}, radiology examinations \cite{bhayana2023performance}, \cite{bhayana2023gpt}, and mathematical tests \cite{floridi2020gpt}. While these studies have demonstrated the potential of GPT models in tackling complex and domain-specific questions, they have also identified challenges and limitations, such as the susceptibility of the models to hallucinate information that is not present in the input data \cite{bender2021dangers}, \cite{wang2021adversarial}.

Contextual information plays a crucial role in the performance of language models, particularly when answering questions that require a deep understanding of specific domains. In this regard, integrating relevant contextual information into the input prompts can significantly enhance the models’ ability to generate accurate and coherent responses \cite{liu2023pre}, \cite{borgeaud2022improving}. However, the optimal strategies for incorporating context and the impact of context length and format on the models’ performance have not been thoroughly investigated.

In this paper, we examine the performance of the GPT-3 model in a question-answering task using contextual information from the California Driver's Handbook. Initially, we investigated the model's performance with no context provided. That is, we tested the GPT-3 model solely on the questions, without utilizing the California Driver's Handbook as a context source. Given that the model's training data likely includes extensive information on driver's knowledge test questions, we specifically wanted to test how well it adapted to the unique rules of the California driver's license knowledge test, as driving regulations can vary slightly from one US state to another and significantly between countries. We explore the influence of varying context lengths and formats on the model’s accuracy and analyze the model’s limitations, including sensitivity to text formatting and potential hallucination. By conducting a series of experiments, we aim to shed light on the factors that contribute to the model’s success and shortcomings in this domain. Our study contributes to the growing body of literature on GPT models in question-answering tasks by offering insights into the role of context in the model’s performance. Moreover, we provide recommendations for optimizing context integration and formatting to enhance the accuracy of the generated responses. These findings have implications for researchers and practitioners employing large-scale language models in various applications, particularly those that involve answering questions based on contextual information.

\section{Methods}
\label{sec:headings}

Figure 1 shows an overview of the proposed method for question answering using contextual information. Specifically, the approach involves retrieving the most relevant parts of the document library (i.e., information source) by means of document embeddings and integrating them into the prompt for GPT. The proposed methodology consists of four major components including: a) Preprocessing of contextual information, b) Embedding of query and identifying the most relevant document sections, c) Prepending of document sections, and d) generating answer using GPT. Each of the four major components will be described in more detail in the following subsections.

\begin{figure}
  \centering
  \includegraphics[width=1\textwidth]{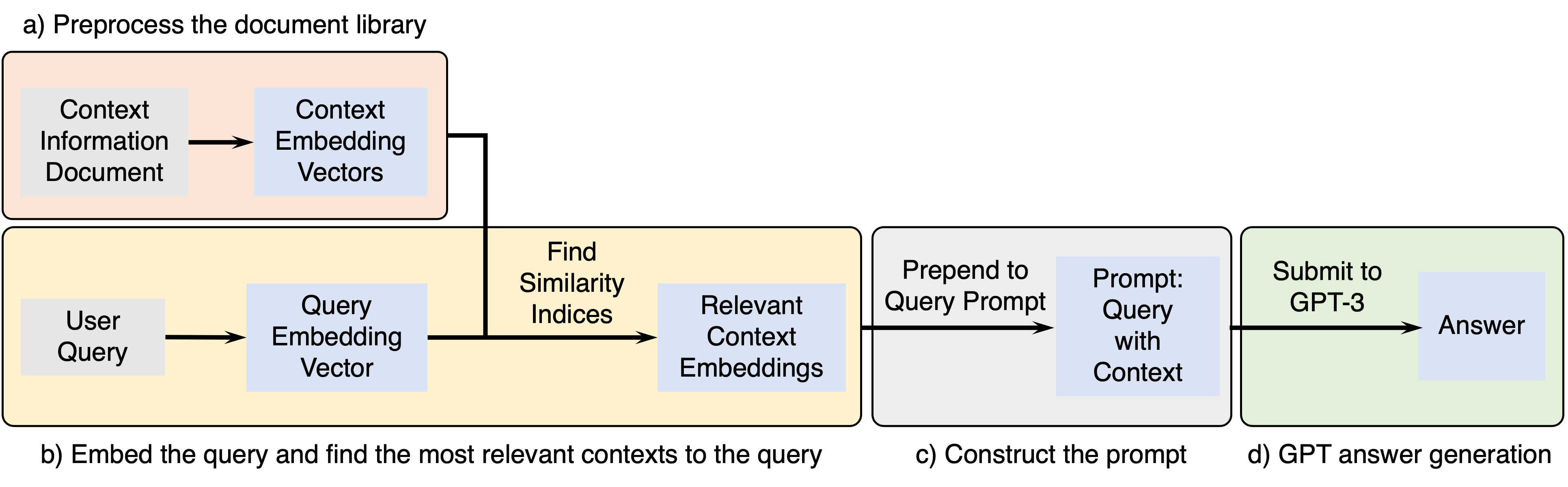}
  \caption{Overview of the proposed method.}
  \label{fig:fig1}
\end{figure}

\subsection{Preprocess the contextual information}
We employed a preprocessing technique to convert contextual information into a numerical representation suitable for further analysis. Initially, we segmented the contextual information into distinct "chunks", each of which could be independently searched and retrieved. Each chunk was then assigned a unique embedding vector, a mathematical representation of its semantic meaning within a high-dimensional space. The distance between two embedding vectors provides a measure of their semantic similarity, a technique frequently employed in various NLP applications, such as sentiment analysis and information retrieval. These document embeddings were stored in a database to facilitate efficient retrieval during query processing.

The California Driver's Handbook was utilized as the primary source of contextual information for this study. To facilitate our analysis, we manually converted the handbook from its original PDF format into a structured tabular dataset, which comprises 108 rows and four distinct columns: "title", "heading", "content", and "tokens". The "title" column represented the 14 main sections of the handbook, the "heading" column denoted subsections, and the "content" column housed the corresponding textual information. Any non-textual components such as images and tables were omitted. Lastly, the "tokens" column indicated the token count of each corresponding content cell. We enforced a token limit for each "content" cell to ensure the length was manageable for the GPT-3 prompt. We aimed to balance the chunk size, making it large enough to contain substantial information yet small enough to accommodate one or several within the GPT-3 prompt. Typically, we found paragraph-length chunks suitable for most applications. Figure 2 illustrates the token distribution within the manually transcribed tabular dataset derived from the California Driver's Handbook.

Subsequently, we utilized Open AI’s text embeddings (embedding model: text-embedding-ada-002)[[9] R. Greene, T. Sanders, L. Weng, and A. Neelakantan, “New and improved embedding model,”
OpenAI, Dec. 15, 2022. https://openai.com/blog/new-and-improved-embedding-model] to numerically represent each "content" cell. The resulting tabular dataset retained the "title" and "heading" columns but replaced the "content" and "tokens" columns with 1536-dimensional embedding vectors. This model, leveraging a transformer-based architecture like BERT and RoBERTa, and a tokenizer named "cl100k-base", can process a maximum of 8191 input tokens, outputting a 1536-dimensional vector for each token. So, we divided our document library into sections and encoded each chunk with its corresponding embedding vector. Tables 1 and 2 respectively present examples of the contextual information and document embeddings data by showcasing three random data rows. In the next phase, we used these document embeddings to facilitate question answering.

\begin{figure}
  \centering
  \includegraphics[width=0.3\textwidth]{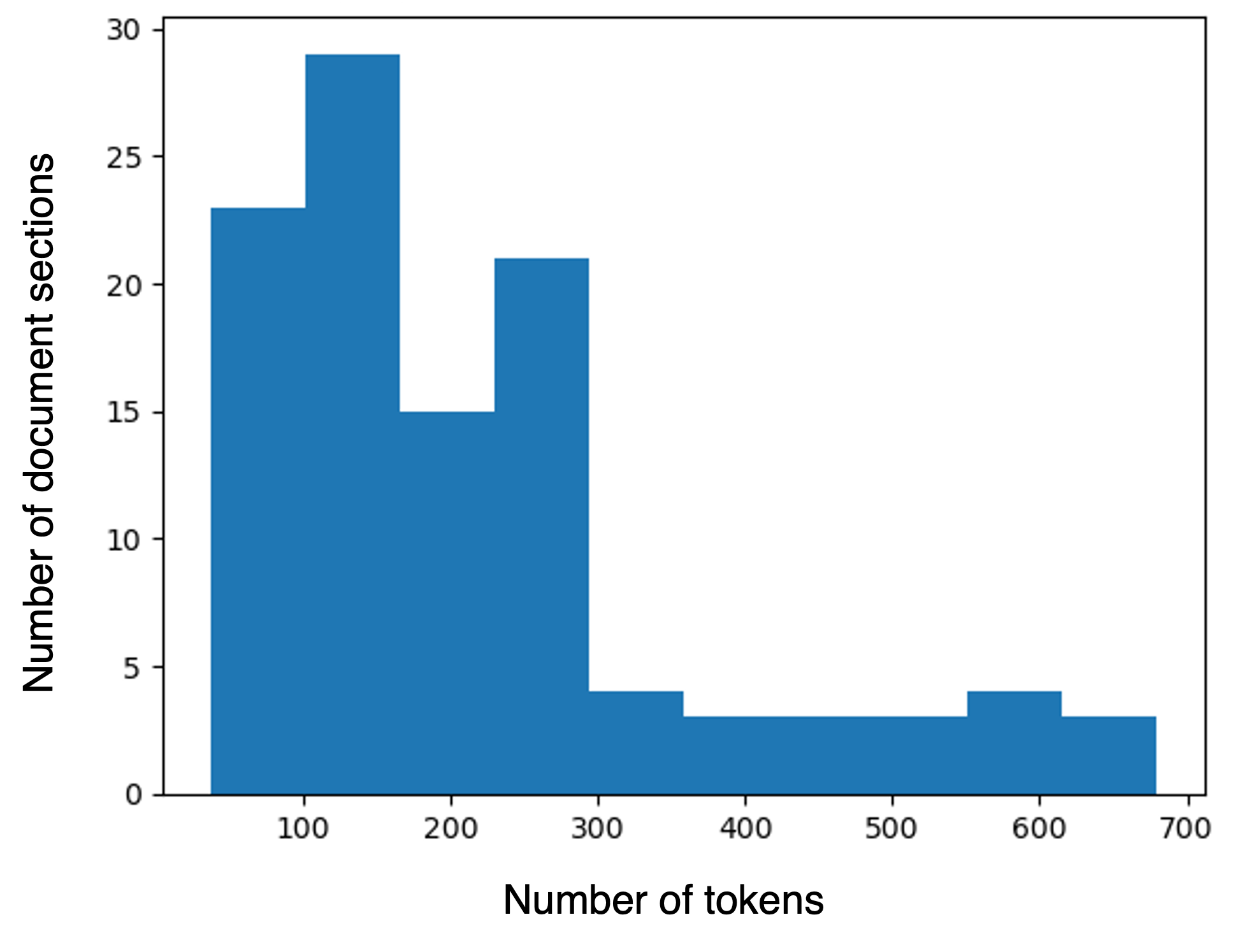}
  \caption{In the tabular dataset created from the California Driver's Handbook, the majority of rows (i.e., content cells) contain fewer than 300 tokens.}
  \label{fig:fig3}
\end{figure}

We then used OpenAI’s text embeddings to convert each “content” cell into a numerical representation (embedding model: text-embedding-ada-002) \cite{WinNT}. The resulting tabular embedding dataset contains the same first two columns (title and heading) where the content and tokens column are replaced by the embedding representation including 1536 float numbers. TEA-002 employs a transformer-based architecture, like BERT and RoBERTa, and utilizes a tokenizer named "cl100k-base". The model can process a maximum of 8191 input tokens and outputs a 1536-dimensional vector representation for each token. So, we have split our document library into sections and encoded them by creating embedding vectors that represent each chunk. Next, we will use these embeddings to answer questions. Table 1 and 2 represent the contextual and embedding data, respectively showing three random rows of data.

\begin{table}[h]
\small
    \caption{Contextual information}
    \centering
    \renewcommand{\arraystretch}{1.5}
    \begin{tabular}{p{2.7cm}p{2.7cm}p{6cm}p{1cm}}
        \toprule
        \textbf{Title} & \textbf{Heading} & \textbf{Context} & \textbf{Tokens}\\
    \midrule
        Vehicle Registration Requirements & Vehicle Registration Requirements & You need to register your vehicle in California in order to use it in the state. For more ... & 204\\

        Changing, Replacing, and Renewing Your Driver's License & Changing Your Information & Change Your Name: If you legally change your name, update your driver's license. Here ... & 194 \\

        The California Driver's License & Who Must Have a Driver's License? & California residents who drive on public roads or use public parking facilities must have a ... & 128 \\
        
    \bottomrule
    \end{tabular}
    \label{tab:info}
\end{table}

\begin{table}[h]
\small
    \caption{Document embeddings}
    \centering
    \renewcommand{\arraystretch}{1.5}
    \setlength{\tabcolsep}{12pt}
    \begin{tabular}{p{2.7cm}p{2.7cm}p{7cm}}
        \toprule
        \textbf{Title} & \textbf{Heading} & \textbf{Embedding Vectors} \\
        \midrule
        Vehicle Registration Requirements & Vehicle Registration Requirements & [0.025, 0.002, 0.034, 0.055, ... , -0.034]\\
        Changing, Replacing, and Renewing Your Driver's License & Changing Your Information & [-0.011, -0.030, 0.004, 0.008, ... ,-0.009] \\
        The California Driver's License & Who Must Have a Driver's License? & [0.022, -0.010, 0.019, 0.053, ... , -0.006]\\
        \bottomrule
    \end{tabular}
    \label{tab:info}
\end{table}

\subsection{Embedding of query and identifying the most relevant document sections}

 Figure 3 provides a comprehensive summary of the steps involved in identifying the most pertinent document sections for a sample question. The first step involves embedding the query within the same vector space as the contextual chunks. This process facilitates the identification of the document embeddings most closely aligned with the query. To do this, we convert the query into a numerical representation, allowing for a direct comparison with the existing document embeddings stored in the database. Upon receiving a query, it undergoes tokenization and is then transformed into a numerical representation. This conversion utilizes the same embedding model previously applied to the contextual information (Figure 3(a)). Following this, the numerical representation of the query is compared to all the document embeddings within the database, in a quest to find the ones bearing the highest similarity to the query. In the context of this study, we employed the sample driver’s license knowledge test obtained from the official site of the California DMV. This sample test is comprised of 50 multiple-choice questions, each offering three potential answers. We had the option of creating an embedding vector solely for the question segment, excluding the multiple-choice answers. However, some questions did not possess sufficient information in their standalone query and required the additional context provided by the multiple-choice answers. As such, for the sake of consistency, we chose to create an embedding vector for the entirety of each question, including the multiple-choice answers. Subsequently, to ascertain the relevance of various sections of the document to a particular query, we used cosine similarity or dot product to measure the similarity between vectors (Figure 3(b)). The similarity indices were then sorted (Figure 3(c)), yielding a list of document sections organized in descending order of relevance (Figure 3(d)). 

\begin{figure}
  \centering
  \includegraphics[width=1\textwidth]{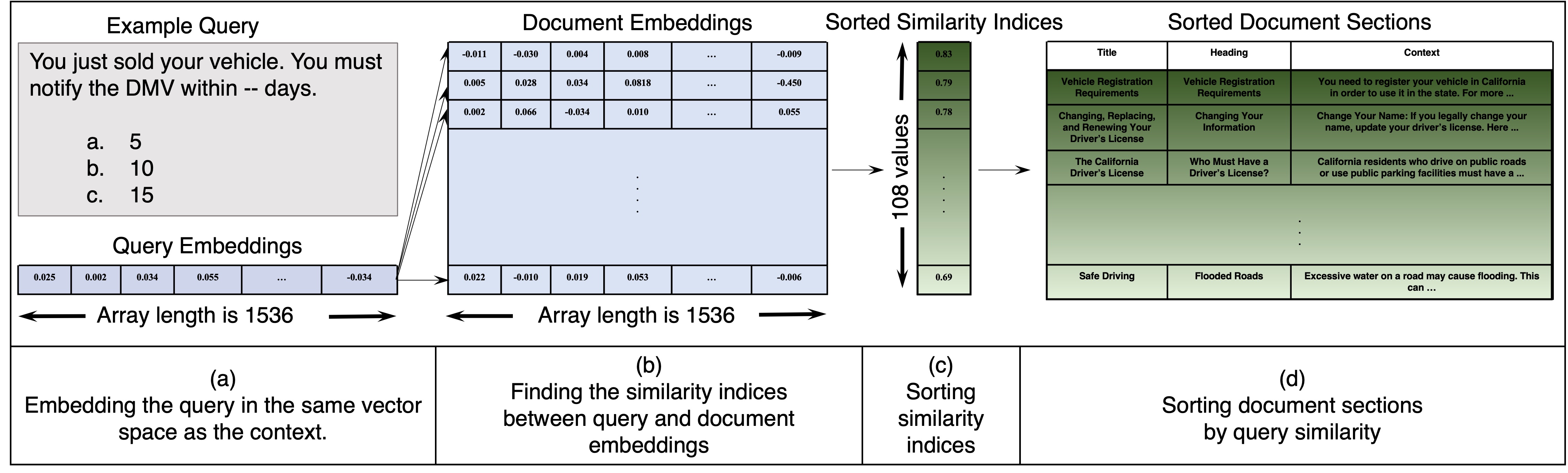}
  \caption{The steps to find the most relevant document sections for a sample question. Finding the query embedding for the supplied query and comparing it against all the pre-calculated document embeddings. Using cosine similarity or dot product to calculate the similarity between vectors to find the most relevant sections. Returning the list of document sections, sorted by relevance in descending order.}
  \label{fig:fig4}
\end{figure}

\subsection{Add the most relevant document sections to the query prompt}

The subsequent step involves appending the most relevant document sections to the query prompt. The number of document sections to be prepended is a challenging task. The length of the prompt or input sequence is an important factor to consider when using GPT models, as longer prompts can provide richer context, thereby generating more precise responses. However, these lengthier inputs may also strain the model and hinder performance speed. The maximum acceptable length for a prompt within GPT models can vary, contingent upon the specific implementation and available hardware resources. For instance, GPT-3, with its 175 billion parameters, can manage prompt lengths up to 2048 tokens. However, in general, it’s a good practice to keep the prompt length as short as possible while still providing enough context for the model to generate relevant responses. This helps ensure optimal performance and avoids overloading the model with unnecessary information.

In our quest to identify the optimal number of document sections to append to the prompt, we conducted experiments with varying prompt lengths given to the GPT model. Specifically, we tested lengths of 500, 1200, and 1900 words. Our findings indicated that a 1900-word prompt length delivered optimal results for most questions. It's important to note that while GPT-3 can handle prompts of up to 2048 tokens, the maximum section length cannot be set at this limit. This is because certain characters must be reserved for the question length and a control statement to prevent the generation of nonsensical responses.  Depending on the length of the user’s question, we found that we could attach a maximum of around 1900 words to the prompt. Upon specifying a certain prompt length, document sections are concatenated at the start of the prompt. Tokens from each document section are added in sequence until the prompt length is reached. This process continues until there is no more space available in the prompt. The GPT model is then fed with the question alongside the most relevant document sections. When a specific prompt length is specified, the document sections are concatenated to the beginning of the prompt, and tokens from each document section are added sequentially until the prompt length is reached. This process continues until there is no more space available in the prompt.

\subsection{GPT-based answer generation}

Having retrieved the relevant context and constructed the query prompt, the next step involves submitting the query, along with the contextually relevant document sections, to the GPT model. The model, having been pre-trained on a vast dataset of natural language processing tasks, employs the provided contextual information to generate a contextually congruent answer. In this procedure, we utilize the OpenAI Completions API to address the user's query via the 'openai.Completion.create()' method. This function, part of the OpenAI API, facilitates the generation of text completions based on the provided prompt or context. It accepts several parameters, such as "engine" (text-davinci-003), "prompt" (the input text), and "temperature" (controlling the randomness of the text). In our work, we set the temperature at 0.0 to ensure the most predictable and factual responses. A low temperature value biases the model towards the most probable outputs, ensuring more predictable and factual responses. Conversely, a high temperature value results in the model opting for less probable outputs, yielding more creative and diverse responses. 

\section{Results}
\label{sec:headings}
\subsection{Engineering prompt length}
This section presents the experimental outcomes of applying the GPT-3 model to a driver's knowledge test. Initially, we investigated the model's performance with no context provided. That is, we tested the GPT-3 model solely on the questions, without utilizing the California Driver's Handbook as a context source. Given that the model's training data likely includes extensive information on driver's knowledge test questions, we specifically wanted to test how well it adapted to the unique rules of the California driver’s license knowledge test, as driving regulations can vary slightly from one US state to another and significantly between countries. Out of the 50 questions tested, GPT-3 model failed to answer 9 questions, resulting in an overall passing score of 82\% (see Table 3). Table 4 presents the details of the 9 questions that GPT-3 model failed to answer, along with the corresponding incorrect answers generated by the model. Next, we conducted experiments with the GPT-3 model when context was provided, in the form of the California Driver's Handbook. This time, the GPT-3 model was tested with incrementally expanding contextual information, beginning at 500 words, then escalating to 1200 and eventually 1900 words (see Table 3). The model's passing score in response to the 500, 1200, and 1900 word contexts were 92\%, 92\%, and 96\% respectively. Interestingly, elevating the word count from 500 to 1200 failed to improve the performance of GPT-3, as it failed at the same 4 questions. Overall, our findings indicate that the GPT-3 model can achieve an 82\% passing score on California driver’s license knowledge test without any context, improving to 96\% when supplemented with context and an adequate word count. However, there are still some questions that the model fails to answer correctly, even with context, indicating that there is still room for improvement in this area. 

\begin{table}[h!]
\small
\label{table:performance_context_prompt_length}
\centering
\caption{The model's passing score in response to no context and with context provided}
\begin{tabular}{cccc}
\toprule
\textnormal{Condition} & \textnormal{Prompt Length} & \textnormal{No. of Failed Questions} & \textnormal{Passing Score} \\\hline
No context provided & N/A & [9, 25, 26, 35, 36, 39, 43, 45, 46] & 82\% \\
Context provided & 500 & [35, 36, 45, 46] & 92\% \\
Context provided & 1200 & [35, 36, 45, 46] & 92\% \\
Context provided & 1900 & [36, 46] & 96\% \\
\bottomrule
\end{tabular}

\end{table}

\begin{table}[h]
\tiny
\centering
\caption{The details of the 9 questions out of the 50 questions that the GPT-3 model failed to answer without/with context, along with the corresponding answers generated by the model.}
\begin{tabular}{|p{0.45cm}|p{6.45cm}|p{1.75cm}|p{1.75cm}|p{1.75cm}|p{1.75cm}|}
\hline
\textbf{\#} & \textbf{Question} & \textbf{GPT3 - No context} & \textbf{500 tokens} & \textbf{1200 tokens} & \textbf{1900 tokens} \\
\hline

Q. 9 & You just sold your vehicle. You must notify the DMV within --- days. \newline
\underline{a. 5} \newline b. 10 \newline c. 15 &
\cellcolor{gray!25} 10 &
5 &
5 &
5 \\
\hline

Q. 25 & You are driving on a freeway posted for 65 MPH. The traffic is traveling at 70 MPH. You may legally drive: \newline
a. 70 mph or faster to keep up with the speed of traffic. \newline
b. Between 65 mph and 70 mph. \newline
\underline{c. No faster than 65 mph.} \newline
& \cellcolor{gray!25}Between 65 mph and 70 mph. 
& No faster than 65 mph. 
& No faster than 65 mph.
& No faster than 65 mph. \\
\hline

Q. 26 & It is illegal to park your vehicle: \newline
\underline{a. In an unmarked crosswalk.} \newline
b. Within three feet of a private driveway. \newline
c. In a bicycle lane. \newline
& \cellcolor{gray!25}In a bicycle lane. 
& In an unmarked crosswalk. 
& In an unmarked crosswalk.
& In an unmarked crosswalk. \\
\hline

Q. 35 & Which of these vehicles must always stop before crossing railroad tracks \newline
\underline{a. Tank trucks marked with hazardous materials placards.} \newline
b. Motor homes or pickup trucks towing a boat trailer. \newline
c. Any vehicle with 3 or more axles or weighing more than 4,000 pounds. \newline
& \cellcolor{gray!25}Any vehicle with 3 or more axles or weighing more than 4,000 pounds.
& \cellcolor{gray!25}Any vehicle with 3 or more axles or weighing more than 4,000 pounds.
& \cellcolor{gray!25}Any vehicle with 3 or more axles or weighing more than 4,000 pounds.
& Tank trucks marked with hazardous materials placards. \\
\hline

Q. 36 & You are driving on a one-way street. You may turn left onto another one-way street only if: \newline
a. A sign permits the turn. \newline
b. Traffic on the street moves to the right. \newline
\underline{c. Traffic on the street moves to the left.} \newline
& \cellcolor{gray!25}A sign permits the turn. 
& \cellcolor{gray!25}A sign permits the turn. 
& \cellcolor{gray!25}A sign permits the turn. 
& \cellcolor{gray!25}A sign permits the turn. \\
\hline

Q. 39 & At intersections, crosswalks, and railroad crossings, you should always: \newline
a. Stop, listen, and proceed cautiously. \newline
\underline{b. Look to the sides of your vehicle to see what is coming.} \newline
c. Slowly pass vehicles that seem to be stopped for no reason. \newline
& \cellcolor{gray!25}Stop, listen, and proceed cautiously. 
& Look to the sides of your vehicle to see what is coming.
& Look to the sides of your vehicle to see what is coming. 
& Look to the sides of your vehicle to see what is coming. \\
\hline

Q. 43 & Always stop before you cross railroad tracks when: \newline
\underline{a. You don’t have room on the other side to completely cross the tracks.} \newline
b. The railroad crossing is located in a city or town that has frequent train traffic. \newline
c. You transport two or more young children in a passenger vehicle. \newline
& \cellcolor{gray!25}The railroad crossing is located in a city or town that has frequent train traffic.
& You do not have room on the other side to completely cross the tracks.
& You do not have room on the other side to completely cross the tracks.
& You do not have room on the other side to completely cross the tracks. \\
\hline

Q. 45 & Should you always drive slower than other traffic? \newline
\underline{a. No, you can block traffic when you drive too slowly.} \newline
b. Yes, it is a good defensive driving technique. \newline
c. Yes, it is always safer than driving faster than other traffic. \newline
& \cellcolor{gray!25}Yes, it is a good defensive driving technique.
& \cellcolor{gray!25}Yes, it is a good defensive driving technique.
& \cellcolor{gray!25}Yes, it is a good defensive driving technique.
& No, you can block traffic when you drive too slowly. \\
\hline

Q. 46 & You see a signal person at a road construction site ahead. You should obey his or her instructions: \newline
a. Only if you see orange cones on the road ahead. \newline
b. Unless they conflict with existing signs, signals, or laws. \newline
\underline{c. At all times.} \newline
& \cellcolor{gray!25}Unless they conflict with existing signs, signals, or laws.        
& \cellcolor{gray!25}Unless they conflict with existing signs, signals, or laws.
& \cellcolor{gray!25}Unless they conflict with existing signs, signals, or laws.
& \cellcolor{gray!25}Unless they conflict with existing signs, signals, or laws.\\
\hline
\end{tabular}
\end{table}

\subsection{Formatting context}

This section presents the findings from our experiment investigating the effect of text formatting on the GPT-3 model's answer generation capabilities. More specifically, we analyzed how alterations in text formatting - transitioning from a format that disregarded bullet points to one that incorporated them - affected the accuracy of the model's responses. As demonstrated in Figure 4, the GPT-3 model generated incorrect answers when the context was delivered without bullet points. In contrast, when the context was formatted with bullet points, the model's output aligned with the correct answer. 

Regarding question 26 (see Table 5), the response for this particular question can be found in the driver’s handbook under the title "Illegal Parking." The original text is presented in a bullet-point format. However, during the preprocessing stage, in which a tabular dataset was created from the handbook, the text was transferred into the "content" cells, and the bullet points were excluded. Instead, a comma separator replaced the bullet points (see Figure 4). Upon further examination of this question to determine the cause of GPT’s inability to provide a satisfactory answer, it was hypothesized that altering the text format to include bullet points could yield better results. When the text was reformatted to consider bullet points, GPT generated an accurate response. Consequently, the embedding vectors for the two scenarios were plotted and compared. Interestingly, it was observed that the embedding vectors for both cases exhibited minor differences. Based on this analysis, it can be concluded that GPT-3 model is sensitive to text formatting.

The same principle applied when considering the question and question embedding. The GPT-3 model seemed to be misled, or "hallucinate", when the three multiple-choice answers were presented in a continuous block of text, leading to an incorrect response. Conversely, when the three answer choices were formatted across three distinct lines, the GPT-3 model demonstrated higher accuracy, correctly selecting only one answer from the provided choices. This experiment underscores the significant role text formatting play in the context provided to the GPT-3 model for answer generation. It's worth noting that to mitigate hallucination effects, we introduced the following statement at the beginning of the prompt: "Use the below text to answer the subsequent multiple-choice question. Pick only one correct answer. If the answer cannot be found, write 'I don't know.'". This method further improved the model's ability to generate accurate responses.

\section{Discussion}
\label{sec:headings}

Considering a prompt length of 1900 tokens and text formatting, GPT-3 model accurately answered 48 questions but was unsuccessful in addressing two questions, specifically Q36 and Q46. An in-depth investigation was conducted to further explore the characteristics of the misleading questions to the model, with the goal of understanding why GPT-3 model failed to generate correct responses for these particular questions. 

\subsection{Context Truncation}

Regarding questions 35 and 45 (see Table 5), the study assessed the influence of varying prompt lengths on the efficacy of the question-answering process by contrasting outcomes when employing prompt lengths of 500, 1200 tokens versus 1900 tokens. The evaluation of question 35 demonstrated that during the identification of the most relevant document sections (see Figure 3), where the most pertinent document sections relevant to the user’s query are filtered, the three document sections from the database with the highest similarity to the query comprised 140, 600, and 524 tokens, respectively. When utilizing a 500 and 1200 -token prompt length, the third document section (containing 524 tokens) was omitted, as the combined length of the first and second sections exceeded 500 tokens. Consequently, the generated prompt only included the initial section, consisting of 140 tokens. In contrast, employing a 1900-token prompt length allowed for the construction of a prompt that incorporated four document sections, with token counts of 140, 600, 524, and 166, respectively. It is essential to note that the answer to the query was situated within the third document section. Thus, expanding the prompt length to 1900 tokens facilitated the inclusion of this section, while the 500 and 1200-token prompts excluded the paragraph containing the answer. This facilitated the generation of an accurate response when the appropriate prompt was submitted to GPT-3. Moreover, extending the prompt length from 500 to 1200 tokens did not yield improved performance, as the supplementary paragraphs were superfluous and devoid of pertinent information. Consequently, an optimal prompt length exists, which permits the inclusion of the query’s answer while preventing the integration of extraneous data. This ultimately enhances the model’s performance and mitigates the occurrence of hallucinations.

\subsection{Confusing question}

Regarding question 36 (see Table 5), it was discovered that the multiple-choice options appeared to be confusing for GPT-3. When the multiple-choice answers were excluded from the question (i.e., "Which of these vehicles must always stop before crossing railroad tracks?" as opposed to "Which of these vehicles must always stop before crossing railroad tracks? a. Tank trucks marked with hazardous materials placards. b. Motor homes or pickup trucks towing a boat trailer. c. Any vehicle with 3 or more axles or weighing more than 4,000 pounds."), GPT-3 provided the response, "Buses, school buses, and trucks transporting hazardous loads," when given contextual information from the driver’s handbook. However, when the same question was posed to GPT-3 without supplying any contextual data, the model generated an incorrect answer. This observation suggests that GPT-3’s performance may be influenced by the complexity of the multiple-choice options provided.

\subsection{Confusing context}

In addressing question 46 (see Table 5), we identified several keywords within the question, which we then sought in the driver’s handbook. These keywords included "signal person," "road construction site," "site ahead," and "obey." Among these, only "obey" appeared in the driver’s handbook, with two instances relevant to the question. Firstly, on page 68, the text states, "Obey special signs or instructions from workers (flaggers)," within the section "Road Workers and Work Zones." Secondly, on page 63, it reads, "Obey any direction, order, or signal given by a traffic officer, law enforcement officer, or firefighter. Follow their orders even if they conflict with existing signs, signals, or laws," found in the "Emergency Vehicles" section. Ideally, when queried, our model should select "Road Workers and Work Zones" as the document section with the highest similarity to the question. However, our model does not appear to establish a connection between "workers (flaggers)" and "You see a signal person at a road construction site ahead." Even if it recognized them as synonymous, there is no mention of "all times" in the handbook. Although it instructs readers to obey special signs or instructions from workers (flaggers), the phrase "all times" is absent. Interestingly, GPT-3’s response to this question is option (b) "Unless they conflict with existing signs, signals, or laws." This hallucination may arise from the exact mention of "conflict with existing signs, signals, or laws" in the context – California Driver’s handbook. Due to the high similarity in wording, GPT-3 may incorrectly judge the question. This observation might indicate that GPT-3 is not conscious, as it merely searches for similarities between contextual information and the query. When encountering the most similar words, it does not employ deep thinking or analysis to make decisions but outputs the most similar multiple-choice option as the correct answer. Our hypothesis suggests that one reason GPT-3 hallucinates is its reliance on word manipulation rather than in-depth question analysis. When it recognizes terms from the text in the multiple-choice options, it selects that answer as correct. To test this, we conducted two experiments. As demonstrated in Table 5, when we exclude the exact words of the text from option "b," the model produces the correct answer, "at all times."

\begin{figure}
\Large
  \centering
  \includegraphics[width=1\textwidth]{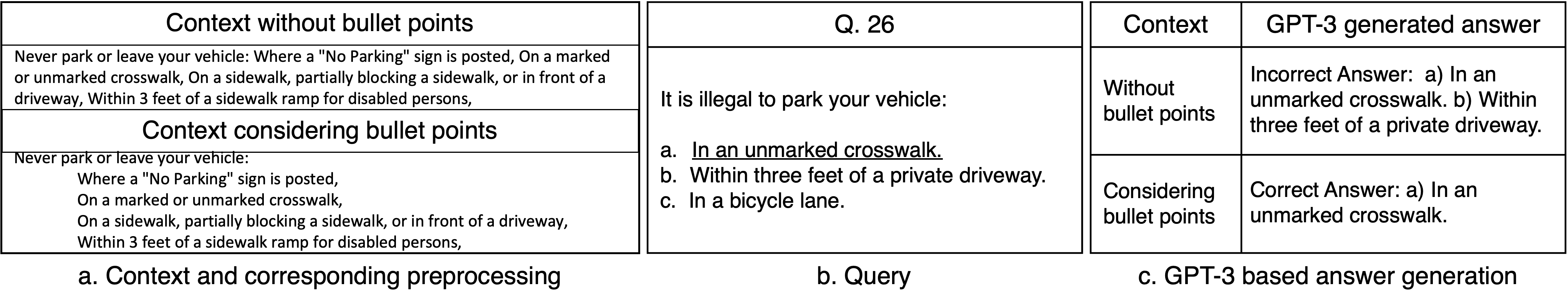}
  \caption{The role of context formatting in the GPT-based answer generation}
  \label{fig:fig5}
\end{figure}

\begin{table}[h!]
\tiny
\centering
\caption{GPT-3 model was unsuccessful in addressing Question 46 when the context was provided.}
\label{table:signal_person_instructions}
\begin{tabular}{|p{8cm}|p{3cm}|p{3cm}|}
\hline
\textbf{Question-46} & \textbf{With general knowledge} & \textbf{With providing the context} \\\hline
You see a signal person at a road construction site ahead. You should obey his or her instructions: \newline a. Only if you see orange cones on the road ahead. \newline b. Unless they conflict with existing signs, signals, or laws. \newline c. At all times. & c) At all times. & b) Unless they conflict with existing signs, signals, or laws.\\\hline
You see a signal person at a road construction site ahead. You should obey his or her instructions: \newline a. Only if you see orange cones on the road ahead. \newline b. Unless they conflict. \newline c. At all times. & c) At all times. & c) At all times.\\\hline
\end{tabular}
\end{table}

\section{Conclusions}

This study has provided valuable insights into the capabilities and limitations of the GPT-3 model for question-answering tasks using contextual information. Our experiments focused on evaluating the model’s performance in answering questions from the California Driver’s Handbook, with and without context, and assessing the impact of context/question format on the generated answers. The results revealed that GPT-3 model achieved an 82\% passing score without context, which improved to 96\% when the context was provided along with a sufficient number of words. However, it is important to note that the model still struggled to answer some questions correctly, even with the appropriate context, suggesting that there is room for improvement in GPT-3 model’s performance for question-answering tasks. In addition, our findings emphasized the significance of context format in enhancing GPT-3’s answer-generation capabilities. This study demonstrates the importance of providing adequate and relevant context to improve the model’s performance in question-answering tasks. Furthermore, it highlights the need to consider context format, such as bullet points, in order to further enhance the model’s ability to generate accurate answers. Finally, it underscores the necessity of identifying and addressing the limitations of GPT-3, such as its sensitivity to text formatting and its potential to hallucinate answers, in order to develop more robust and reliable question-answering systems. Future research in this area could focus on devising strategies to improve GPT-3’s handling of text formatting, as well as exploring other factors that may contribute to its performance in question-answering tasks. Additionally, investigations into the capabilities and limitations of newer or alternative language models, and comparisons with GPT-3, would be valuable for advancing the field of natural language processing and question-answering systems.

\section*{Acknowledgements}

We are grateful to Dr. Zhen Zeng for her fruitful comments and  corrections. 

\bibliographystyle{plain}
\bibliography{references}

\end{document}